\documentclass[letterpaper]{article} 
\usepackage{aaai25}  
\usepackage{times}  
\usepackage{helvet}  
\usepackage{courier}  
\usepackage[hyphens]{url}  
\usepackage{graphicx} 
\usepackage{amsmath}
\usepackage{subfig}
\usepackage{subfloat}
\usepackage{natbib}  
\usepackage{caption} 
\usepackage{amsthm}
\usepackage{amsfonts}

\usepackage{newfloat}
\DeclareCaptionStyle{ruled}{labelfont=normalfont,labelsep=colon,strut=off} 
\usepackage[linesnumbered,ruled,vlined]{algorithm2e}

\title{FedCross: Intertemporal Federated Learning Under Evolutionary Games}
\author {
    Jianfeng Lu\textsuperscript{\rm 1,\rm 2},
    Ying Zhang\textsuperscript{\rm 1},
    Riheng Jia\textsuperscript{\rm 3}$^*$,
    Shuqin Cao\textsuperscript{\rm 1},
    Jing Liu\textsuperscript{\rm 1},
    Hao Fu\textsuperscript{\rm 4}
}

\affiliations {
    \textsuperscript{\rm 1}School of Computer Science and Technology, Wuhan University of Science and Technology, China\\
    \textsuperscript{\rm 2}Key Laboratory of Social Computing and Cognitive Intelligence (Dalian University of Technology), Ministry of Education, China.\\
    \textsuperscript{\rm 3}School of Computer Science and Technology, Zhejiang Normal University, China\\
    \textsuperscript{\rm 4}Hubei Province Key Laboratory of Intelligent Information Processing and Real-time Industrial System, Wuhan University of Science and Technology, China\\
    \{jianfenglu, ravenb7\}@wust.edu.cn, rihengjia@zjnu.edu.cn, \{shuqincao, luijing\_cs, fuhao\}@wust.edu.cn
}

\usepackage{bibentry}

\begin{document}

\maketitle

\begin{abstract}

Federated Learning (FL) mitigates privacy leakage in decentralized machine learning by allowing multiple clients to train collaboratively locally.  However, dynamic mobile networks with high mobility, intermittent connectivity, and bandwidth limitation severely hinder model updates to the cloud server. Although previous studies have typically addressed user mobility issue through task reassignment or predictive modeling, frequent migrations may result in high communication overhead. Addressing this challenge involves not only dealing with resource constraints, but also finding ways to mitigate the challenges posed by user migrations. We therefore propose an intertemporal incentive framework, FedCross, which ensures the continuity of FL tasks by migrating interrupted training tasks to feasible mobile devices. Specifically, FedCross comprises two distinct stages. In Stage 1, we address the task allocation problem across regions under resource constraints by employing a multi-objective migration algorithm to quantify the optimal task receivers. Moreover, we adopt evolutionary game theory to capture the dynamic decision-making of users, forecasting the evolution of user proportions across different regions to mitigate frequent migrations. In Stage 2, we utilize a procurement auction mechanism to allocate rewards among base stations, ensuring that those providing high-quality models receive optimal compensation. This approach incentivizes sustained user participation, thereby ensuring the overall feasibility of FedCross. Finally, experimental results validate the theoretical soundness of FedCross and demonstrate its significant reduction in communication overhead.
\end{abstract}

\section{Introduction}

With the explosive growth of artificial intelligence (AI), concerns regarding privacy and security issues associated with centralized learning have been intensifying \cite{survey-privacy}. Federated Learning (FL), introduced by Google \cite{Communication-efficient}, marks a paradigm shift in distributed learning, tackling data privacy and communication challenges by decentralizing model training. FL mitigates the risk of private data leakage by allowing multiple clients to perform local training based on a globally shared model, and then sending the updated model parameters to a central server for aggregation. However, frequent communication between clients and servers can lead to significant overhead. Using edge servers, such as base stations, as intermediaries for early aggregation in a hierarchical federated learning (HFL) framework \cite{Adaptive-hfl} is undoubtedly a better choice. Since users participating in model training inevitably consume computational and communication resources \cite{LOSP}, it is essential to design an effective HFL framework that strategically incentivizes users to contribute high-quality model updates during the training.

Currently, research on incentives in HFL may still be insufficiently comprehensive. On one hand, in FL, mobile users within a region may not be able to consistently provide services to the current base station, as the number of users in a specific area may not remain constant over an extended period \cite{cross-area}. As the migration rate increases, the model accuracy is expected to decline and the convergence speed will decrease, while the associated communication overhead will also increase. Unfortunately, previous research has largely focused on single issues \cite{IOT-based-migration}, neglecting the interplay between task migration decisions and resource allocation schemes. Task migration is a factor affecting task continuity, while resource allocation influences task migration decisions. On the other hand, although existing studies have designed appropriate incentives to encourage users to participate in training, the sustainability of the incentives versus the limited budget is again a problem \cite{fairness-aware}. Therefore, it is necessary to implement appropriate incentive mechanisms to mitigate these negative impacts.

Addressing the above issues is highly urgent and most confront three major challenges. (i) \textit{Ensuring task continuity during migration.} Some participants may leave the coverage area of the current base station at any given moment, leading to interruptions in FL tasks \cite{Joint-allocation}. Consequently, maintaining task continuity while minimizing migration overhead is a critical challenge to prevent data wastage. (ii) \textit{Non-IID data distribution among mobile users.} As numerous users participate in the training process, different data distribution among mobile users arising from diverse local environments may lead to a loss of model accuracy \cite{FedCDA}. (iii) \textit{Incentivizing user participation over time.} With resource constraints and limited user rationality, participants may lose interest or be infeasible to continue contributing over a longer period of time. Therefore we need to ensure the sustainability of the incentives.

Considering the aforementioned issues and challenges, we introduce a spatiotemporal Hierarchical Federated Learning framework (FedCross) that leverages evolutionary game theory and procurement auctions to ensure the continuity of FL tasks during dynamic training. Specifically, we first simulate task migration through binary crossover and polynomial mutation operations among individuals, and perform online task allocation based on each user's communication resources. FedCross innovatively integrates evolutionary game theory to construct a dynamic clustering model. By analyzing the stability of the replicator dynamic equations using replicator dynamic functions, we maximize user participation rates. Additionally, FedCross employs procurement auctions to simulate the trading process, ensuring that base stations providing high-quality model updates receive maximum benefits. Overall, FedCross is an effective solution for addressing user mobility in FL, particularly in highly dynamic networks such as vehicular networks. The main contributions of this paper are as follows:

\begin{itemize}
\item[$\bullet$] The FedCross framework is proposed to address the online incentive problem for mobile users, considering both users' bounded rationality and resource limitations. It leads to effective task migration strategies and resource allocation schemes, ensuring continuity of FL task training while improving model accuracy under limited communication and computational resources.

\item[$\bullet$] The dynamic decision-making process of mobile users is modeled as an evolutionary game. A cross-region online mobility algorithm is proposed to allocate resources based on users' channel capacity, addressing the user mobility challenge in FL training.

\item[$\bullet$] To further incentivize user participation, the transaction between cloud servers and base stations is modeled as a greed-based procurement auction. The optimal set of winners is explored, ensuring fairness for both parties.

\item[$\bullet$] Numerical simulations were conducted to support our theoretical analysis and verify the validity of the proposed framework.
	
\end{itemize}

\section{Related Work}
\textbf{Research on Mobility in FL. }Due to the characteristics of mobile users and the geographical limitations of regionally deployed base stations, cross-region issues are inevitable in current FL research. However, privacy constraints, such as those outlined by the GDPR \cite{GDPR}, strictly prohibit research involving users' actual geographical locations. Traditional solutions to this problem typically involve transfer learning, where a model trained in one region is applied to another \cite{Transfer-Learning}. Nevertheless, in the FL scenario, the heterogeneity of user data across different regions can lead to significant model bias when transferring models. Some studies have further explored this issue. For instance, \citeauthor{Enhancing-FL}\shortcite{Enhancing-FL} segmented the global model in heterogeneous FL scenarios and considered factors such as edge devices' storage capacity and network conditions for migration. In real-world healthcare scenarios, \citeauthor{LossAware}\shortcite{LossAware} applied FL for patient data transfer, effectively reducing the average total cost using the minimum cost algorithm (MCA) in FedAvg. Notably, these studies are not applicable in dynamically changing user mobility scenarios. To address this, we propose FedCross to capture users' dynamic cross-region behavior.

\textbf{Game-Based Incentive Mechanisms in FL}. Many researchers tend to utilize game theory as the foundation for incentive design, thereby ensuring the effectiveness of the mechanism. \citeauthor{Economic-stackelberg}\shortcite{Economic-stackelberg} employed a Stackelberg game model to optimize FL in wireless environments, where the incentive is realized by deriving the optimal reward function for the base station. \citeauthor{FL-Zero-Determinant}\shortcite{FL-Zero-Determinant} modeled the interaction between the server and devices during FL training as a repeated game, using a zero-determinant strategy-based incentive mechanism to achieve incentivization. Similarly, \citeauthor{Optimality-stability}\shortcite{Optimality-stability} uses coalition game theory to calculate the optimal arrangement of users, discussing user incentives by providing the Price of Anarchy bounds between individual incentives and social welfare. While \citeauthor{When-fl-meets}\shortcite{When-fl-meets} proposes a novel framework called FedGame, which constructs a non-cooperative game to incentivize MEC nodes to actively participate in resource competition in the face of IIoT attacks. However, the aforementioned games are not suitable for scenarios where users change their decisions over time. Therefore, some works have turned their attention to evolutionary games. For instance, \citeauthor{Cloud-federation-formation}\shortcite{Cloud-federation-formation} leverages evolutionary game theory to maintain the stability of coalition members in the presence of dynamic strategies. However, their centralized mechanism might undermine the independence of user decisions. In contrast, FedCross adopts evolutionary games to predict dynamic decisions across user regions, thereby mitigating frequent user migration behaviors.

\textbf{Auction in Federated Learning}. In FL (auction-based mechanisms), auction theory is typically a crucial tool for incentive mechanism design and game-theoretic analysis. For instance, in the satellite edge cloud domain, \citeauthor{Incentive-Double-Auction}\shortcite{Incentive-Double-Auction} proposed a quantity-dependent double auction incentive mechanism for participating satellites, aiming to achieve efficient resource allocation and fair incentives for participating nodes. \citeauthor{Privaim}\shortcite{Privaim} introduced a multi-dimensional reverse auction-based double privacy-preserving incentive mechanism, which addresses the issue of sensitive data leakage through bid obfuscation and task reassignment. \citeauthor{Large-scale}\shortcite{Large-scale} proposed an efficient large-scale personalizable bidding method for multi-agent auction-based federated learning. Inspired by these works, we adopt a procurement auction to ensure the identification of the maximum number of optimal winners under communication constraints, thereby achieving long-term incentives for users.

\section{System Model}
\subsection{System Setting}
In this section, we provide the architecture of FedCross with corresponding details.

\textbf{Structure of FedCross:} In our comprehensive framework, FedCross, the cloud server responsible for aggregating model parameters is denoted by $\mathcal{L}$. Additionally, there exists a set of Base Stations (BS), represented as $\mathcal{B_s} = \left\{1, \cdots, b_s, \cdots, B_s\right\}$
, which are located in various regions with good mobile network coverage. Within these corresponding regions are mobile users, denoted by $\mathcal{N}=\left\{1,\cdots,n,\cdots,N\right\}$. Without loss of generality, we assume that each user ($n\in\mathcal{N}$) can only establish a connection with one BS ($b_s\in\mathcal{B_s}$) and uses their smart devices to interactively train a globally shared FL model using their local datasets.

\textbf{Workflow of FedCross:} As illustrated in Fig.\ref{fig:FL_workflow}, each round of FL training in FedCross consists of four stages:

\begin{figure}[t]
	\centering
	\includegraphics[width=0.45\textwidth]{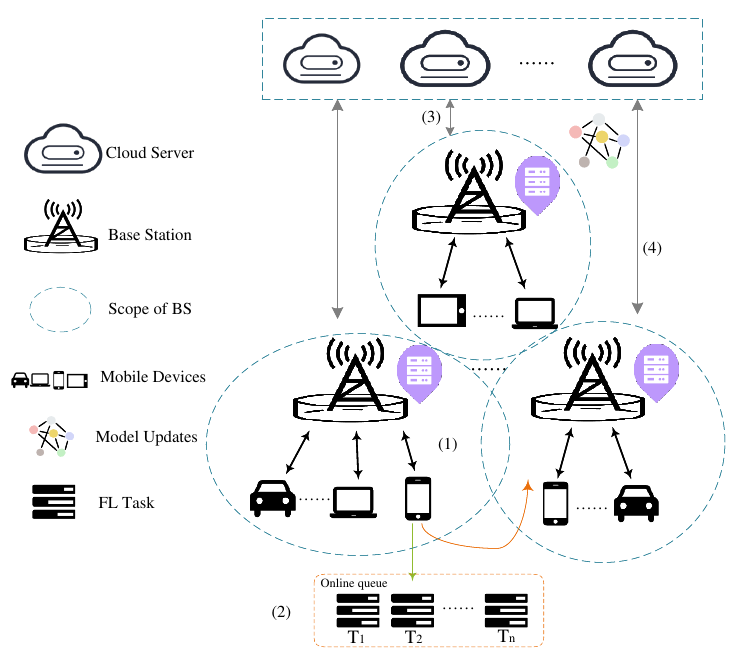}
	\caption{The workflow of FedCross framework.}
	\label{fig:FL_workflow}
\end{figure}

(1) \textit{Formation of Regions:} The process of region formation is conceptualized as an evolutionary game, wherein users make decisions based on the rewards offered by BSs and their respective mobility strategies. Through multiple iterations of evolution, optimal strategies gradually emerge, leading to a stable clustering of users.

(2) \textit{Mobility-Based Local Training via Migration Algorithm:} Users participate in the training of the global model using their local data. The tasks of users leaving the region are recorded in an online queue, where receiving users are selected based on the resource requirements of these tasks for intra-region migration. The aggregation of updated gradient parameters is handled by the BS, which then transmits the merged parameters to the cloud server for integration.

(3) \textit{Greedy-Based Procurement Auction:} The cloud server collects local model accuracy and bids which provided by the BSs and employs a greedy algorithm to select the winning combination. Subsequently, the actual payment required from the winning BSs is determined based on a critical value rule.

(4) \textit{Aggregation and Distribution:} The cloud server then distributes the global model to the winning BSs, which in turn disseminate it to the members within their regions for iterative training on mobile devices. This process is repeated until the training is completed.

\subsection{Communication Model}

During the training process of FL tasks, communication between mobile users and the base station typically involves two components: the base station propagates the global model, and users upload gradient updates. Given that the transmission power of the base station is significantly higher than that of user devices \cite{Massive-MIMO, Scheduling-Aggregation}, we do not consider channel allocation behavior and compression errors in the downlink in this paper. However, users face resource constraints in their upload rates\cite{FedDSE}. Therefore, it is necessary to design an efficient task allocation scheme for mobile users to mitigate these adverse effects.

We consider a block fading channel model, where the channel coefficients remain constant during each FL iteration. Simultaneously, we employ Orthogonal Frequency Division Multiple Access (OFDMA) \cite{OFDMA} to ensure orthogonal resource allocation across different links, thereby preventing interference during users' upload of updated parameters. Both large-scale fading $\beta_k$ and small-scale fading $h_n(t)$ are considered for the channels between the $n$-th mobile users and BSs. Let the transmit power of the $n$-th user's device be $P_n(t) \leq P_n^{max}$, and the power of the additive white Gaussian noise (AWGN) be $\epsilon \sim \mathcal{N}(0, \sigma_w^2)$. According to the Shannon formula \cite{Shannon-formula}, the channel capacity for the $n$-th user is given as follows:

\begin{equation}
Q_n(t) = \log_2 \left( 1 + \frac{P_n(t) \beta_n |h_n(t)|^2}{\sigma_w^2} \right),
\label{eq:channel_capacity}
\end{equation}

\noindent where $\beta_n$ and $h_n$ denotes the large-scale and small-scale fading of the channel from the n-th mobile device to the BS, respectively.

Furthermore, to reduce communication overhead and enhance the training efficiency of each participant, we propose that user devices employ an appropriate model shifting compression scheme \cite{SoteriaFL} based on the channel capacity given by (\ref{eq:channel_capacity}). For privacy and security considerations, we allow users to add Gaussian noise to the gradient updates, expressed as $g_t^n = \tilde{g}_t^n + \xi_t^n$, where $g_t^n$ is the gradient shift, and $\xi_t^n$ represents noise drawn from a normal distribution $N(0,\sigma_p^2I)$. This perturbation reduces the model’s reliance on individual users and makes data recovery more challenging for potential attackers. During each iteration, participants compress their gradient estimates using a compression operator $C:\mathbb{R}^d \rightarrow \mathbb{R}^d$, resulting in a compressed vector $v_t^i = C(\tilde{g}_t^i)$, which is then sent to the base station. Finally, the cloud server aggregates all compressed vectors to update the global model.

\subsection{Online Migrate Strategies For Cross Areas}

\textbf{Trigger migration. }During local model training, the task status of mobile users may change. When they move out of the current area, the BS is unable to receive model parameters due to geographical constraints. To ensure the continuity of FL training and further enhance model accuracy, 
inspired by \cite{Workflow-migration}, we employ a resource-constrained migration algorithm to facilitate cross-regional training for mobile users. For users with bounded rationality, we posit that they will trigger the migration process when they cannot achieve reasonable returns in the current area.

In such cases, users preparing to leave will terminate their tasks prematurely and send updated parameters to the online queue to obtain prior training rewards. However, due to the early termination, the training quality is relatively lower, resulting in the offline users receiving fewer rewards. Nonetheless, they still receive some compensation to ensure their efforts are not wasted. The online queue will poll and execute the migration algorithm, dynamically reallocating the interrupted tasks to other users within the current area to ensure that data is not wasted.

\textbf{Migration process. }Algorithm \ref{alg:OMC} illustrates the task migration process within the online queue. Our migration strategy is built upon the foundation of a genetic algorithm. In each iteration, two individuals are randomly selected from the population $P$, and based on their dominance relationship, one is added to the mating pool $P_t^m$. Each individual in the mating pool then undergoes crossover and mutation operations, generating new offspring. The parent and offspring individuals are combined and undergo non-dominated sorting. Through environmental selection, the non-dominated fronts that meet the size constraints are added to the next generation population $P_{t+1}$. Subsequently, task migration is performed based on the available resources of each user. We calculate the channel capacity $Q_n(t)$ for each user and decide on task assignment based on this capacity. If a user's channel capacity meets the task requirements, the task is assigned to that user. The computational overhead in this part mainly comes from the non-dominated sorting step, which has a complexity of $O(N^2)$ per iteration for $N$ devices. To address this, we employ parallel execution of selection, crossover, and mutation during the migration phase, and experimental results confirm the effectiveness of this optimization. Finally, $P_{t+1}$ is set as the new population $P$ for the next iteration. Upon reaching the maximum number of iterations $t_{\text{max}}$, the final population $P_{t_{\text{max}}}$ is returned as the result.

\section{Theoretical Analysis of FedCross}
\subsection{Stage 1: Evolutionary Games}
We model the dynamic behavior of users moving in different areas as an evolutionary game\cite{fl-evolutionary-game} as follows:

\textbf{Definition 1, }\textit{An evolutionary game for mobile user selection can be described as a four-tuple ($\mathcal{N}, \mathcal{S}, \mathbf{X}, U$), where $\mathcal{N}$ is the set of users, $\mathcal{S}$ is the strategy space, $\mathbf{X}$ is the population state, and $\mathcal{U}$ is the utility function.}

At the initial stage, users randomly enter a region based on their needs, forming an initial population. In the evolutionary game model, each mobile user dynamically adjusts their strategy according to their utility. Throughout the strategy evolution process, users with bounded rationality continuously undergo trial and error, gradually changing their strategies over time. They decide whether to continue uploading model parameters to the base station in the current region or migrate the FL task to stop losses in a timely manner by calculating the mobile payoffs.

\textbf{Utility Function and Replicator Dynamics.} The base station determines the reward based on the model accuracy uploaded by the users, which in turn affects a portion of the users' utility. The strategy adaptation process and the corresponding training strategy evolution can be modeled and analyzed using replicator dynamics \cite{Dynamics-IOT-Evolutionary}, which are a set of ordinary differential equations.

We use $u_{b_s}$ to represent the utility obtained by the user from the BS $b_s$, i.e.,

\begin{equation}
    u_{b_s}(x_b) = R_{n,b_s} \frac{x_{n,b_s} M_n}{\sum_{b_s=1}^{B_s} x_{n,b_s} M_n} - \xi Q_n,
\end{equation}

\noindent where ${R_{n,b_s}}$ represents the reward allocation held by BS $b_s$, $M_n$ denotes the data volume of user $n$, $Q_n$ denotes the channel capacity as previously mentioned, and $\frac{x_{n,b_s} M_n}{\sum_{b_s=1}^{B_s} x_{n,b_s} M_n}$ represents the reward share based on the data contribution of the mobile user, while $\xi$ is the per-unit training cost. Additionally, we assume that $U(\cdot)$ is a linear utility function representing the risk neutrality of data owners without loss of generality \cite{UAV-risk-neutrality}. The net utility that users in the area of BS $b_s$ can obtain at time $t$ is given as follows:

\begin{algorithm}[tb]
    \caption{Online Migrate Strategies For Cross Areas}
    \label{alg:OMC}

    \KwIn{Initial population $P = \{P_1, P_2, \ldots, P_N\}$, Online task queue $T$}
    \KwOut{Final population $P_{t_{\text{max}}}$}

    \For{$t = 1$ \KwTo $t_{\text{max}}$}{
        $P_t^m = \emptyset$
        \For{$i = 1$ \KwTo $N$}{
            $x_1 = \text{RandomSelect}(P)$
            $x_2 = \text{RandomSelect}(P)$
            \If{$x_1$ dominates $x_2$}{
                $P_t^m = P_t^m \cup \{x_1\}$
            }
            \Else{
                $P_t^m = P_t^m \cup \{x_2\}$
            }
        }

        $Q_t^m = \emptyset$
        \For{individual $i \in P_t^m$}{
            $child = \text{SBX}(P_t^m[i])$
            $mutant = \text{PM}(child)$
            $Q_t^m = Q_t^m \cup \{mutant\}$
        }

        $Z = P \cup Q_t^m$

        $P_{t+1} = \emptyset$
        $F = \text{NonDominatedSorting}(Z)$
        \For{each front $F_h \in F$}{
            \If{size of $F_h \leq N - |P_{t+1}|$}{
                $P_{t+1} = P_{t+1} \cup F_h$
            }
        }

        \For{each task $T_j \in T$}{
            \For{each user $u \in P_{t+1}$}{
                \If{$C_k$ (based on Eq.\ref{eq:channel_capacity}) is sufficient for $T_j$}{
                    Assign $T_j$ to user $u$
                    \textbf{break}
                }
            }
        }

        $P = P_{t+1}$
    }

\end{algorithm}

\begin{equation}
    u_{b_s}(x_{b_s}(t)) = R_{n, b_s} \frac{x_{n, b_s}(t) \mathcal{M}_n}{\sum_{b_s=1}^{B_s} x_{n, b_s}(t) \mathcal{M}_n} - \xi Q_n(t).
\end{equation}

Accordingly, the average utility of the area with user \( b \) as the BS is:

\begin{equation}
    \overline{u}(x_{b_s})(t) = \sum\limits_{b_s=1}^{B_s} u_{b_s}(x_{b_s})(t) x_{b_s}(t).
\end{equation}

After multiple generations of evolution, a steady-state strategy, known as an Evolutionarily Stable Strategy (ESS)\cite{Pricing-spectrum}, will emerge as the solution. This is a more successful strategy that propagates over time. When the entire system reaches an ESS state, the population proportions of different species will be in a stable condition. Correspondingly, the replicator dynamic equation is given by:

\begin{equation}
    \dot{x}_{b_s}(t) = y_{b_s}(x(t)) = \Delta x_{b_s}(t)(u_{b_s} - \overline{u}), \forall \, b_s \in \mathcal{B}_s,
    \label{eq:replicator_dynamic}
\end{equation}

\noindent where $\Delta$ refers to the learning rate of strategy adaptation.

\textbf{Theoretical Analysis of Replicator Dynamics. }We provide a theoretical analysis of the evolution of the proportion of the population in a region as a result of user mobility.

\textbf{Lemma 1} \textit{The first-order derivatives of \( y_{b_s}(x(t)) \) with respect to \( x_{\hat{b_s}}(t) \) are bounded for all \( \hat{b_s} \in B_s \).}

\textbf{Theorem 1 }For any initial condition \( x(0) \in X \), there exists a unique evolutionary equilibrium to the dynamics defined in Equation (\ref{eq:replicator_dynamic}).

\textbf{Theorem 2 }For any initial condition \( x(0) \in X \), the evolutionary equilibrium to the dynamics defined in Equation (\ref{eq:replicator_dynamic}) is stable.

Consequently, we demonstrate that within FedCross, areas modeled under evolutionary game theory can achieve dynamic stability. Even if users cross different areas for varied factors, FedCross can continually replicate successful strategies, leading back to a stable state.

\subsection{Stage 2: Greedy-Based Procurement Auction}
\textbf{Auction Design. }Inspired by \cite{incentive-auction,cellular-auction}, we propose a greedy-based procurement auction to model the allocation between BSs and cloud servers. Existing works has explored solutions to the allocation problem, such as deep learning-based auctions \cite{Decentralized-Edge-Intelligence} and greedy policy auctions \cite{Fmore}. Unlike the above, the scenarios in this paper take into account the transaction process between BSs and servers in a cross-regional environment, realising the incentives for the participants. 

\textbf{Allocation Rule. }First, we need to determine the set of winning base stations. To ensure fairness, we require that at least $K$ base stations are selected in each round, and each base station can be selected at most once per round. Based on the training within regions from the previous phase, we take into account the communication overhead in each region. Regions with lower resource consumption are prioritized, as this indicates higher training efficiency of the users within the region \cite{cost-effictive}.

The model quality obtained by mobile users training within the region and the valuation cost of the base station together form the bid price $\text{Bid}_{b_s, j}$. Thus, our allocation rules can be formalized as follows:

\begin{equation}
    \label{eq:base_determine}
    \begin{aligned}
        & \min \sum_{b_s \in \mathcal{B_s}} \sum_{j \in J} \text{Bid}_{b_s, j} \cdot x_{b_s, j} \\
        & \text{s.t.} \quad \left\{
        \begin{aligned}
            & \sum_{b_s \in \mathcal{B_s}} y_{b_s}(t) \geq K, \quad \forall j \in J, \\
            & T_g \geq \frac{1}{1 - \max_{b_s,j} \text{Accur}_{b_s, j}} x_{b_s,j}, \\
            & \sum_{j \in J} x_{b_s,j} \leq 1, \quad \forall b_s \in \mathcal{B_s}, \\
            & x_{b_s,j} \leq \frac{t_{\text{cmp}} + Q_n(t)/\eta}{t_{\text{max}}^{b_s}}. \\
        \end{aligned}
        \right.
    \end{aligned}
\end{equation}

For the aforementioned social cost minimization problem, we solve it using a greedy algorithm \cite{resource-auction}. Here, $T_g$ is the global iteration count calculated based on the maximum model accuracy provided by the selected base stations, $\text{Accur}$ is the model accuracy provided by the $j$-th bid of base station $b_s$, and $Q_n(t)$ is the channel capacity as previously described. 
 $y_{b_s}(t)$ is a binary variable that indicates whether the BS $b_s$ is selected during the global iteration $ t$. Specifically, $ y_{b_s}(t) = 1 $ if client$ i $ is selected in iteration$t $; otherwise, $ y_{b_s}(t) = 0$.

\textbf{Payment Rule. }To ensure individual rationality and incentive compatibility of the auction mechanism, we calculate the payments based on the critical bid using the critical value rule \cite{Truthful-mechanisms}. We determine the payment for each newly selected schedule by computing it relative to the critical bid. The process of finding the critical bid involves first identifying the bid with the second smallest average cost among all feasible bids and then calculating the payment for each selected schedule based on the critical value rule. Then, the algorithm design of the whole auction is shown as follows:

Our algorithm \ref{alg:OptimizedBaseStationSelection} addresses the base station selection problem under multiple constraints eq.(\ref{eq:base_determine}) using a greedy strategy to minimize the payment cost. We initialize an empty set $S$ to store selected base stations and iterate until the number of selected base stations meets or exceeds the predetermined minimum $K$. In each iteration, we select a base station-task combination $(b_s^, j^)$ from the bidding set $J^{T_g}$ that satisfies all constraints and has the lowest cost. Once identified, the combination is added to $B_S$ and removed from $J^{T_g}$ to prevent duplicates. In the payment rule section, we compute the payment for each selected base station by finding its critical bid. Finally, we return the selected base stations $B_S$ and their payments $P$.

The complexity analysis reveals that the main complexity arises from the search for the lowest-cost base station that meets the constraints. In the worst case, this phase has complexity $O(K \times n)$, where $K$ is the minimum number of base stations and $n$ is the size of the bidding set. Each iteration requires $O(n)$ to find the lowest-cost base station, and up to $K$ base stations must be selected, leading to a total complexity of $O(K \times n)$. The payment rule phase also has complexity $O(K \times n)$, as finding the critical bid for each base station takes $O(n)$ time. Therefore, the overall time complexity is $O(K \times n)$, meaning the runtime is linearly dependent on $K$ and $n$.

\begin{algorithm}[t]
    \caption{Optimized Base Station Selection with Payment Calculation}
    \label{alg:OptimizedBaseStationSelection}
    \KwIn{Set of qualified bids $J^{T_g}$, utility increment $R_{il}(S)$, number of global iterations $T_g$, minimum number of base stations $K$, communication time $t_{\text{cmp}}$, channel capacity $Q_n(t)$, maximum allowable time $t_{\text{max}}^{b_s}$}
    \KwOut{Set of selected base stations $B_S$ and payments $P$}
    
    Initialize an empty set $S$ for selected base stations\;
    
    \While{$|S| < K$}{
        Find the bid $(b_s^*, j^*)$ in $J^{T_g}$ with the minimum cost satisfying:
        \[
        T_g \geq \frac{1}{1 - \text{Accur}_{b_s^*, j^*}} \quad \text{and} \quad \frac{t_{\text{cmp}} + Q_n(t) / \eta}{t_{\text{max}}^{b_s^*}} \geq 1
        \]
        Add base station $b_s^*$ to the set $S$\;
        Set $x_{b_s^* j^*} \gets 1$\;
        
        Remove $(b_s^*, j^*)$ from $J^{T_g}$\;
    }
    
    \ForEach{$b_s$ in $B_S$}{
        Find the critical bid $(b_s, j_0)$ such that:
        \[
        (b_s, j_0) = \arg\min_{(b_s, j) \in J^{T_g} \setminus S} \left\{ \frac{r_{b_s, j}}{R_{b_s j}(S)} \right\}
        \]
        Calculate the payment for the selected bid $(b_s^*, j^*)$ using the critical value rule:
        \[
        p_{b_s^*} = R_{b_s^* j^*}(S) - \frac{r_{b_s j_0}}{R_{b_s j_0}(S)} R_{b_s^* j^*}(S)
        \]
        Set the payment $p_{b_s^*}$ for the base station $b_s^*$\;
    }
    
    \KwRet{Set of selected base stations $B_S$ and payments $P$\;}
    
\end{algorithm}

\textbf{Property Analysis. }Based on the above, we analyze the properties of FedCross by the following theorems.

\textbf{Theorem 1.} \textit{The payment rule satisfies both individual rationality and incentive compatibility. Specifically, each bidder's payment ensures that their utility is non-negative, i.e., $v_{b_s} \geq \theta_{b_s}$, and a bidder with a bogus bid $\widetilde{Bid_{b_s}} \neq v_{b_s}$ does not gain more utility compared to bidding truthfully $Bid_{b_s} = v_{b_s}$.}

\begin{table}[t]
    \centering
    \caption{Simulation Parameters}
    \label{tb-table1}
    \setlength{\arrayrulewidth}{1.2pt} 
    \begin{tabular}{c|c}
        \hline  
        \rule{0pt}{8pt} 
        Parameters & Simulated Values \\
        \hline  
        Total number of Servers & 10 \\
        Total number of Areas & [2, 3] \\
        Total number of users & [50, 300] \\
        Congestion coefficient & 10 \\
        Reward & [600, 900] \\
        Momentum & (0, 0.9) \\
        \hline  
    \end{tabular}
    \setlength{\arrayrulewidth}{0.4pt} 
\end{table}

The above theorems ensure that, regardless of the behavior of other BSs, honest bidding remains the optimal strategy for each BS in the auction. All theoretical proofs in the paper are available in the appendix.

\section{Experiment}

\subsection{Experimental Setups}

\textbf{Baselines. }We compared with the following methods:

\begin{itemize}
    \item[$\bullet$] \textbf{BasicFL:} This baseline overlooks user migration and incentive mechanisms, representing a basic FL framework similar to \cite{BasicFL}, where training assumes an ideal environment.

    \item[$\bullet$] \textbf{SAVFL:} \citeauthor{SAVFL}\shortcite{SAVFL} uses simulated annealing to optimize virtual machine migration, focusing on the optimal path and strategy. However, it failed to address frequent migrations, unlike the FedCross.

    \item[$\bullet$] \textbf{WCNFL:} \citeauthor{WCNFL}\shortcite{WCNFL} Introduces a reverse auction incentive mechanism, allowing service providers to select cost-effective devices within budget, enhancing global model performance and user participation.
\end{itemize}

\textbf{Implementation. }We use PyTorch \cite{pytorch} to implement FedCross and the other baselines. The other data is shown in the below table \ref{tb-table1}.

\subsection{Experimental Results}

\begin{figure*}[t]
    \centering
    \subfloat[The proportion state of areas.]
    {
        \includegraphics[width=0.24\linewidth, height=0.20\linewidth]{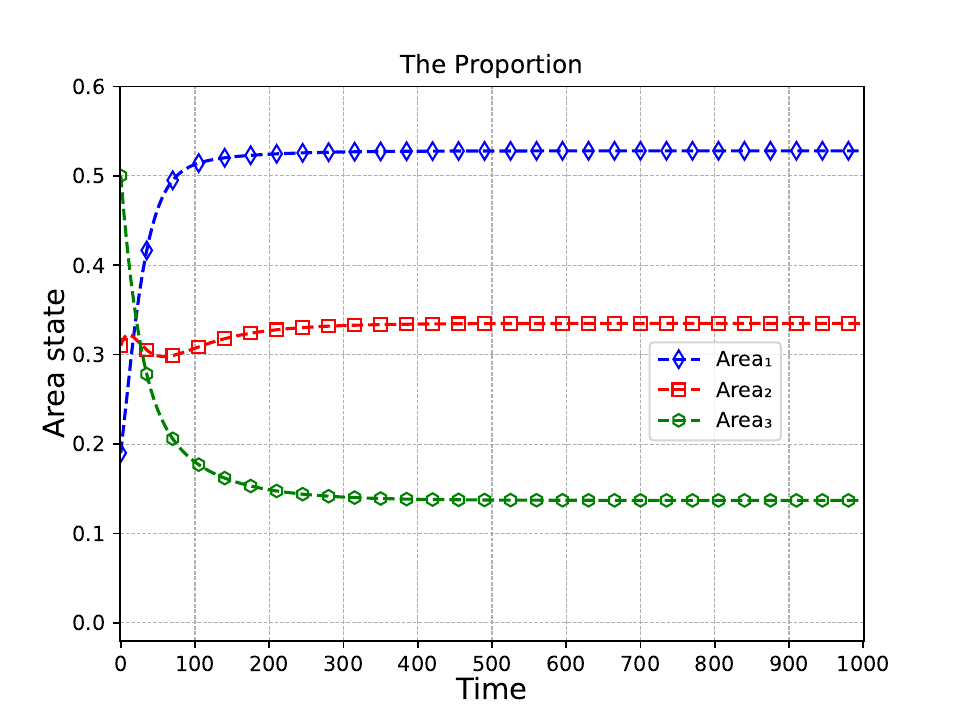}
        \label{fig:Area_3}
    }
    \subfloat[Stability in 3D space.]
    {
        \includegraphics[width=0.24\linewidth, height=0.20\linewidth]{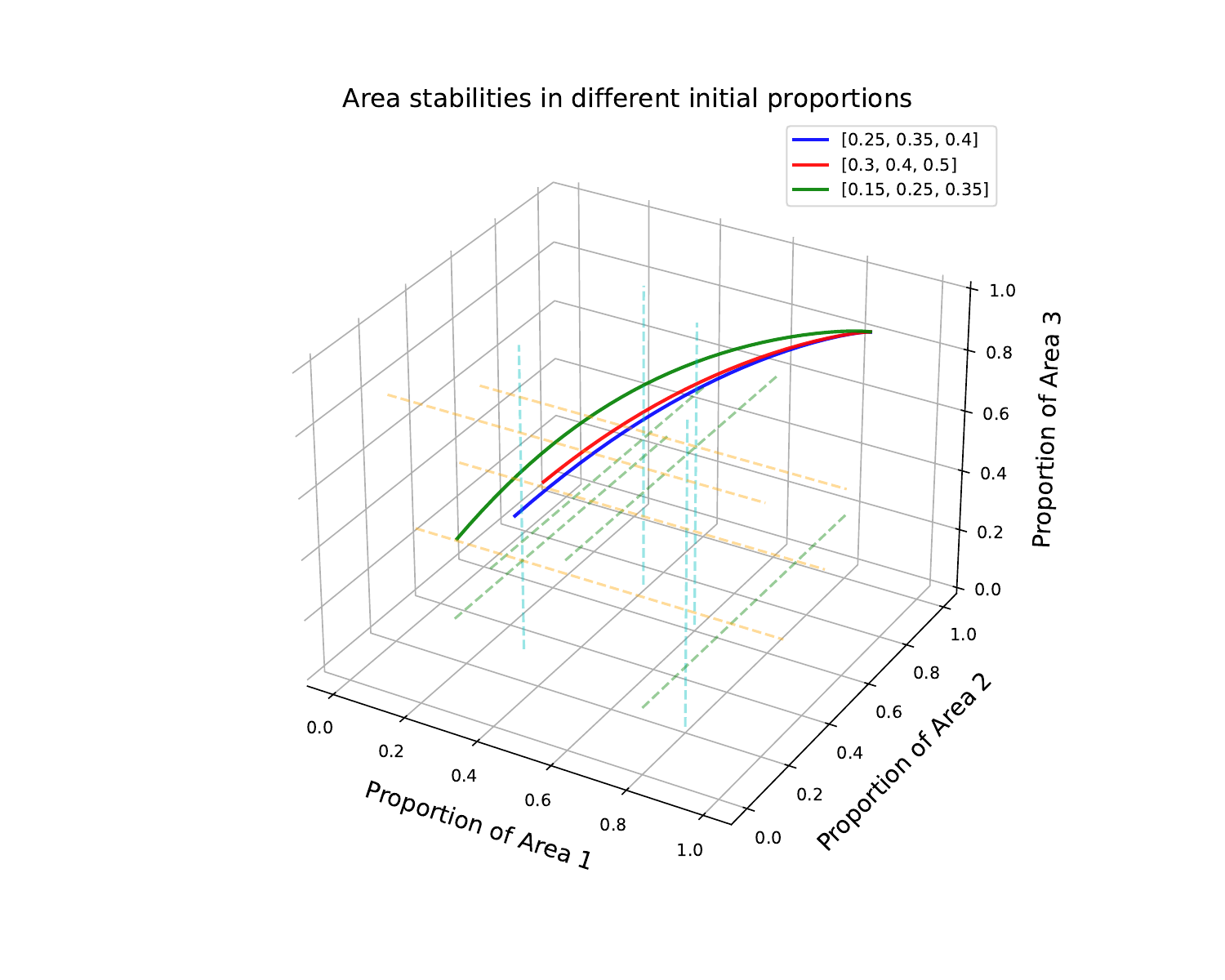}
        \label{fig:Evo_3D}
    }
    \subfloat[The migration situation.]
    {
        \includegraphics[width=0.24\linewidth, height=0.20\linewidth]{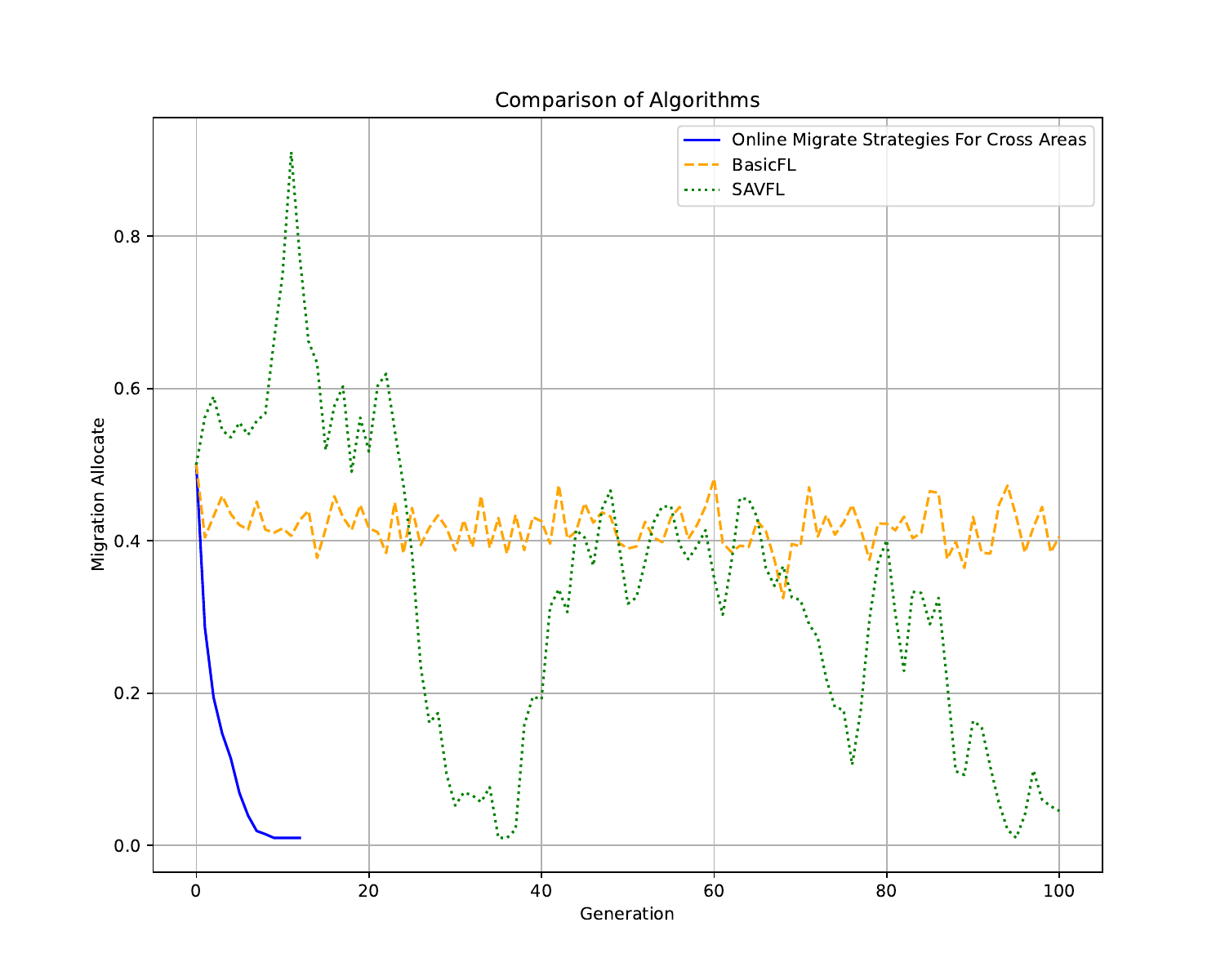}
        \label{fig:Migrate}
    }
    \subfloat[Incentive for users.]
    {
        \includegraphics[width=0.24\linewidth, height=0.20\linewidth]{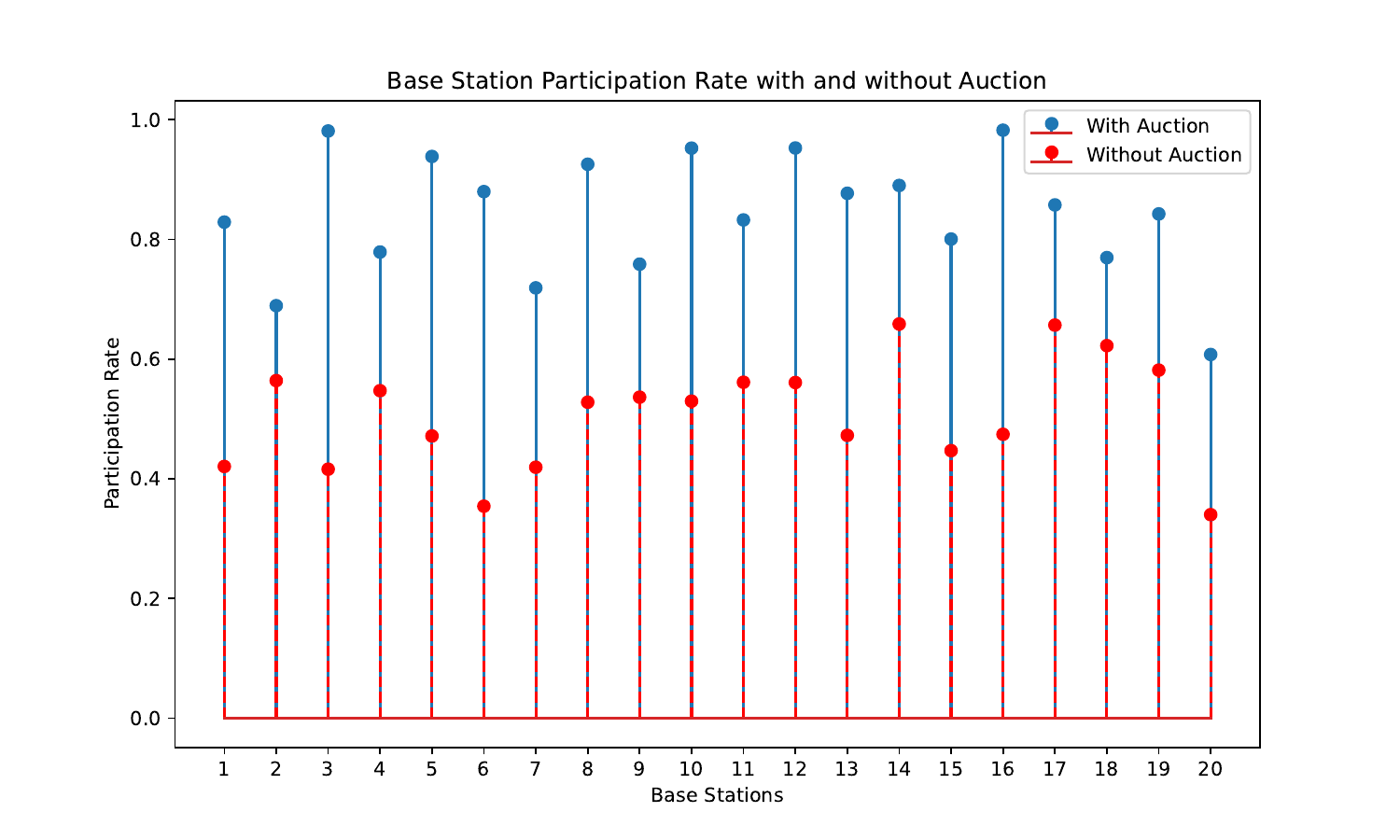}
        \label{fig:BS_Participation_Rate}
    }
    \caption{FedCross under different baselines.}
    \label{fig:Fig_all}
\end{figure*}

\begin{figure}[t]
    \centering
    \subfloat[User-paid costs.]
    {
        \includegraphics[width=0.49\linewidth, height=0.38\linewidth]{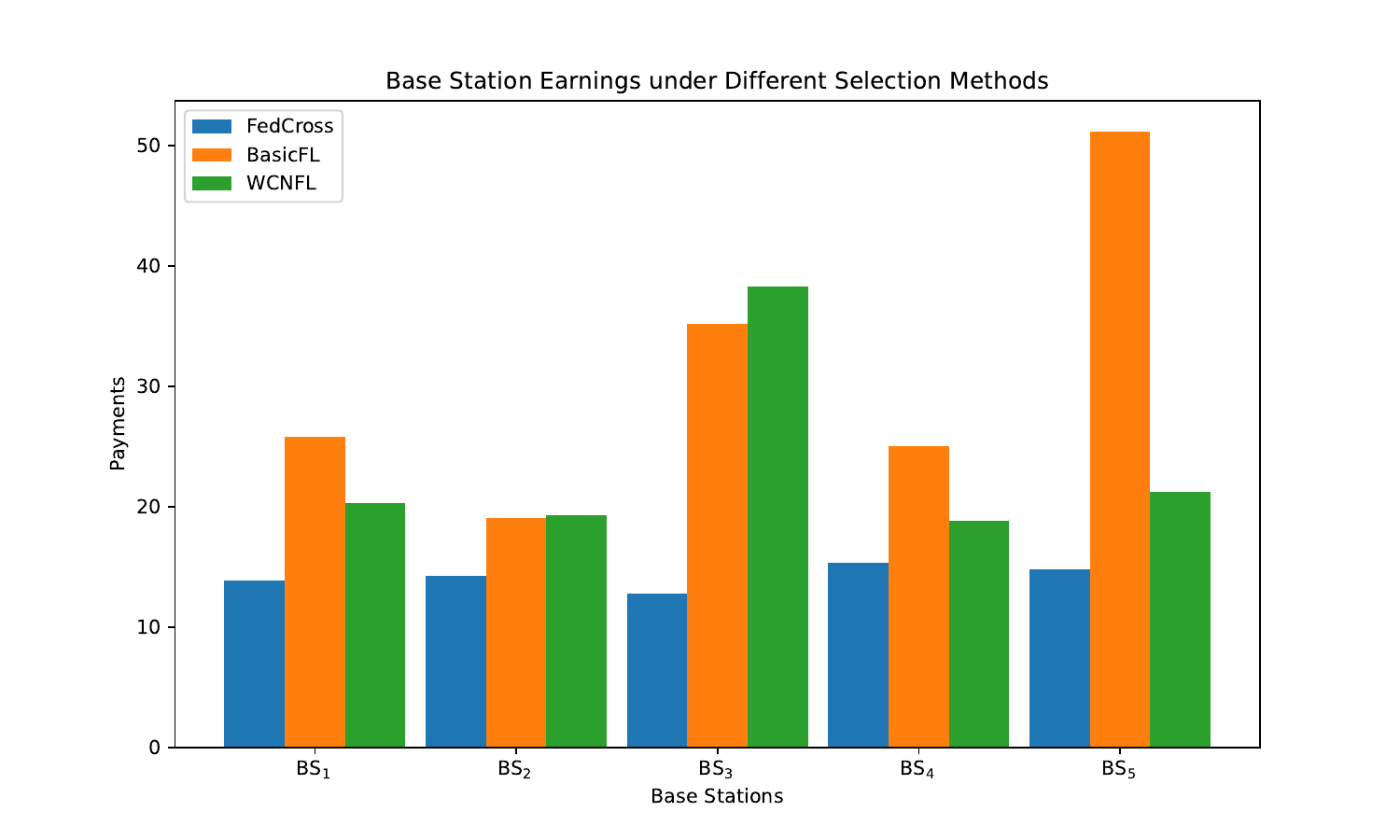}
        \label{fig:BS_Gain}
    }
    \subfloat[Impact of Payment.]
    {
        \includegraphics[width=0.49\linewidth, height=0.38\linewidth]{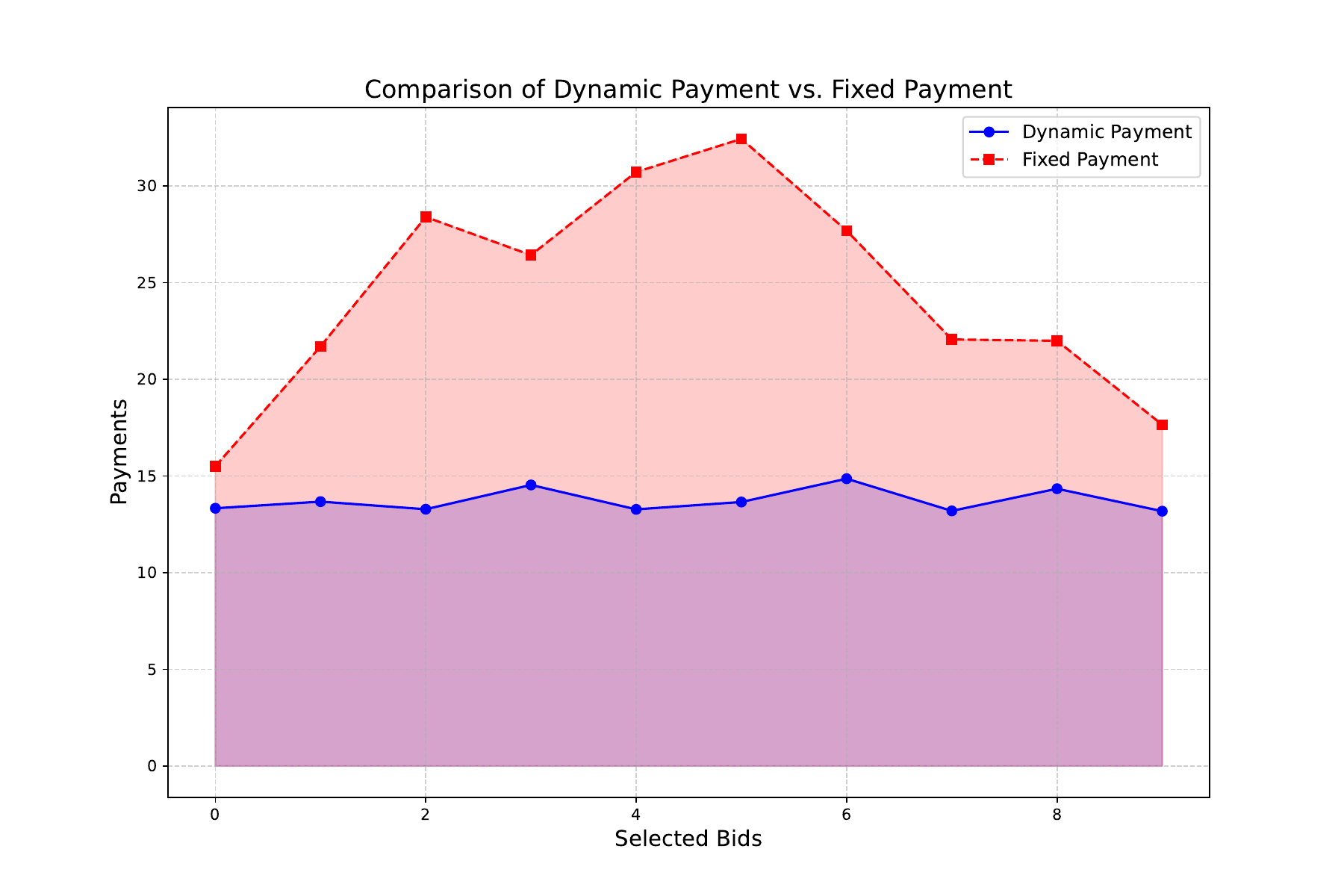}
        \label{fig:Payment}
    }
    \caption{Auction impact.}
    \label{fig:Fig_all_2}
\end{figure}

We provide the population proportion evolution resulting from regional user migration strategies under different scenarios, demonstrate the applicability of user migration processes under a multi-objective evolutionary algorithm, and assess the role of the auction algorithm in the base station selection process, thus validating the theoretical correctness.

\textbf{Evolution of migration. }In Fig.\ref{fig:Area_3}, we first present the overall distribution of the three areas in a two-dimensional state. Given that dynamic changes may occur throughout the training process, we selected a specific time segment for analysis to demonstrate that the proposed framework can achieve dynamic equilibrium under theoretical eq.(\ref{eq:replicator_dynamic}) support. When the user distribution proportions in the three areas are 18$\%$, 32$\%$, and 50$\%$, respectively, the blue curve exhibits an upward trend. This indicates that when mobile users migrate and enter this area, the current base station can attract more users to continue their training in this area. It is important to emphasize that this explanation does not imply that a large number of users choose to train in this area. Such an assumption would be unreasonable, as users' migration decisions are influenced by various external factors, including geographical location and reward levels. Instead, this proportion refers to the percentage of migrating users who are willing to continue their training in the current area. This phenomenon can reduce the frequency of users exiting the training in the current area, thereby lowering the communication overhead in frequent migrations.

In contrast, the green curve shows a significant decline, indicating that the incentives and migration compensation provided by the current base station fail to offset the communication costs, leading users to abandon further training in this area. The red curve initially rises slightly, then drops rapidly, followed by a slight increase, and eventually stabilizes. This suggests that dynamic stability has been reached within this area. After time exceeds 300, the vertical axis reveals that the training proportions across the areas tend to stabilize. Again, this does not imply that users cease migrating. Stability here is a dynamic concept, designed to demonstrate the game-theoretic equilibrium we aim to establish.

In reference to Fig.\ref{fig:Evo_3D}, it further substantiates our previous explanation based on Fig.\ref{fig:Area_3}. The three areas corresponding to the blue curve have proportions of [0.25, 0.35, 0.4], while the red curve corresponds to [0.3, 0.4, 0.5], and the green curve corresponds to [0.15, 0.25, 0.35]. It is evident that as time progresses, the system's final state converges consistently (though this does not imply that the user proportions within each area are identical).

\textbf{Migration process under algorithms. }In Fig.\ref{fig:Migrate}, we present the task allocation results under different baselines. Specifically, BasicFL represents an ideal scenario that neglects user mobility, and therefore employs random search as a simulation method. It can be observed that, due to the lack of a clear optimization direction, the yellow curve fails to significantly improve the task allocation performance, resulting in a relatively poor overall outcome. In contrast, SAVFL explores different solutions in the initial phase; however, due to the early exploration of suboptimal solutions, its initial performance is subpar. As the solution space gradually narrows, the algorithm converges and ultimately achieves superior performance, although this convergence requires a significant number of iterations. Finally, the migration algorithm we propose, by comprehensively considering multiple objectives such as resource overhead and fairness loss, is able to quickly update and return a more optimal task allocation scheme with fewer iterations.

\textbf{Incentive effects of auctions. }Fig.\ref{fig:BS_Participation_Rate} and Fig. \ref{fig:BS_Gain}, we illustrate the role of the auction in the second phase of our framework. Fig. \ref{fig:BS_Participation_Rate} demonstrates the incentivization effect of the auction on user participation, with an increased number of base stations to more clearly reveal this effect. As shown in the figure, users are more inclined to participate in training when tangible rewards are offered by the base stations. In Fig. \ref{fig:BS_Gain}, we compare the payment costs incurred under different allocation rules during the auction phase. BasicFL, which follows a traditional auction allocation rule, results in higher payment costs. WCNFL employs a reverse auction mechanism to select winners; however, its optimization effect on eq.\ref{eq:base_determine} is still inferior to that of FedCross.

In Fig.\ref{fig:Payment}, we demonstrate the effectiveness of the threshold-based payment algorithm. In contrast, although the non-payment algorithm (depicted by the red curve) occasionally shows a downward trend under the same payment costs, the overall user payment cost exhibits unstable growth. This high degree of fluctuation may lead to a decline in user trust, as the rewards provided by the base station are highly unpredictable, making it difficult to anticipate future reward amounts. On the other hand, the payment algorithm (represented by the blue curve) exhibits a more stable and lower payment cost, which not only helps maintain consistent user participation but also ensures the accuracy of the models provided by the base station, thereby positively impacting the overall system performance.

\section{Conclusion}

Our paper introduced FedCross, a spatiotemporal incentive framework that addresses FL resource allocation and task continuity in mobile user scenarios, while ensuring improved model accuracy through incentivized user participation. FedCross employs a multi-objective migration algorithm for continuous task training, uses evolutionary game theory to model dynamic user decisions, and leverages procurement auctions to reward base stations for high-quality updates, encouraging sustained participation. Finally, experimental results demonstrate its effectiveness.

\section{Acknowledgements}
This work was supported in part by the National Natural Science Foundation of China under Grants 62372343, 62272417, 62402352, and 62072411, in part by the Zhejiang Provincial Natural Science Foundation of China under Grant LR21F020001, in part by the Key Research and Development Program of Hubei Province under Grant 2023BEB024, and in part by the Open Fund of Key Laboratory of Social Computing and Cognitive Intelligence (Dalian University of Technology), Ministry of Education under Grant SCCI2024TB02.

\appendix


\newpage

\appendix
\section{Appendix}
\subsection*{Proof of Lemma 1}

Consider the replicator dynamic function:

\begin{equation}
\dot{x}_{b_s}(t) = y_{b_s}(x(t)) = \Delta x_{b_s}(t)(u_{b_s} - \overline{u}),
\end{equation}
we take the first-order derivative of \( y_{b_s}(x(t)) \) with respect to \( x_{\hat{b_s}}(t) \):

\begin{equation}
\begin{split}
\frac{\partial y_{b_s}(x(t))}{\partial x_{\hat{b_s}}(t)} 
= \ & \Delta \left( \frac{\partial x_{b_s}(t)}{\partial x_{\hat{b_s}}(t)} (u_{b_s} - \overline{u}) \right. \\
& \left. + \ x_{b_s}(t) \frac{\partial (u_{b_s} - \overline{u})}{\partial x_{\hat{b_s}}(t)} \right).
\end{split}
\end{equation}

Since \( x_{b_s}(t) \) represents a proportion, it is naturally bounded within [0,1]. The utility function \( u_{b_s} \) is given by:

\begin{equation}
    u_{b_s}(x_b) = R_{n,b_s} \frac{x_{n,b_s} M_n}{\sum_{b_s=1}^{B_s} x_{n,b_s} M_n} - \xi Q_n,
\end{equation}

The term \( \frac{\partial (u_{b_s} - \overline{u})}{\partial x_{\hat{b_s}}(t)} \) remains bounded due to the structure of the utility function, where \( \sum_{b_s=1}^{B_s}x_{n,b_s}M_n \) ensures that the denominator is non-zero and bounded.

Thus, we conclude that \( \frac{\partial y_{b_s}(x(t))}{\partial x_{\hat{b_s}}(t)} \) is bounded, proving the lemma.

\subsection*{Proof of Theorem 1}
\subsubsection*{Stage 1}
Starting with Theorem 1, we note that the replicator dynamics, given by:
\begin{equation}
y_{b_s}(x(t)) = \Delta x_{b_s}(t)(u_{b_s} - \overline{u}),
\end{equation}
is a continuous and differentiable function. From Lemma 1, we have established that the first-order derivatives of this function are bounded, implying the system satisfies the Lipschitz condition.

Applying the Mean Value Theorem to \( y_{b_s}(x(t)) \), we find that there is a constant \( c \) such that:
\begin{equation}
y_{b_s}(x_1(t)) - y_{b_s}(x_2(t)) = \frac{dy_{b_s}(c)}{dx_{\hat{b_s}}} (x_1(t) - x_2(t)),
\end{equation}
the boundedness of \( \frac{dy_{b_s}(c)}{dx_{\hat{b_s}}} \) ensures the uniqueness of the solution, thus proving the existence and uniqueness of the equilibrium.

\subsubsection*{Stage 2}
To prove Theorem 2, we consider the Lyapunov function:
\begin{equation}
G(x(t)) = \sum_{b_s=1}^{B_s} x_{b_s}(t)^2.
\end{equation}

Taking its derivative with respect to time, we have:
\begin{equation}
\frac{dG(x(t))}{dt} = 2 \sum_{b_s=1}^{B_s} x_{b_s}(t) \dot{x}_{b_s}(t).
\end{equation}

Substituting \( \dot{x}_{b_s}(t) = \Delta x_{b_s}(t)(u_{b_s} - \overline{u}) \) into the equation, we get:
\begin{equation}
\frac{dG(x(t))}{dt} = 2 \Delta \sum_{b_s=1}^{B_s} x_{b_s}(t) x_{b_s}(t)(u_{b_s} - \overline{u}),
\end{equation}
at equilibrium, where \( \sum_{b_s=1}^{B_s} \dot{x}_{b_s}(t) = 0 \), the derivative \( \frac{dG(x(t))}{dt} \) equals zero, satisfying the conditions for Lyapunov stability. Hence, the equilibrium is stable.

\subsection*{Proof of Theorem 2}
\subsubsection*{Stage 2}
IR requires that each BS's utility—defined as the difference between their valuation and payment—remains non-negative. In the payment calculation of the algorithm, for each selected base station \( b_s^* \), the payment \( p_{b_s^*} \) is determined using a critical value rule:
\begin{equation}
p_{b_s^*} = R_{b_s^* j^*}(S) - \frac{r_{b_s j_0}}{R_{b_s j_0}(S)} R_{b_s^* j^*}(S).
\end{equation}

Here, \( R_{b_s^* j^*}(S) \) represents the utility increment brought by the selected base station \( b_s^* \), and \( \frac{r_{b_s j_0}}{R_{b_s j_0}(S)} \) indicates the utility increment of other bidders in the absence of \( b_s^* \). Since the payment \( p_{b_s^*} \) will not exceed the actual utility increment, the utility for the BS is:
\begin{equation}
u(b_s^*) = R_{b_s^* j^*}(S) - p_{b_s^*} \geq 0,
\end{equation}
which ensures non-negative utility for each bidder, satisfying IR.

IC ensures that bidders maximize their utility by bidding truthfully. If a BS \( b_s \) submits a false bid \( \widetilde{Bid_{b_s}} \neq v_{b_s} \), the payment is recalculated based on this bogus bid. The payment depends on the actual utility increment \( R_{b_s^* j^*}(S) \) and the critical value \( \frac{r_{b_s j_0}}{R_{b_s j_0}(S)} \). Misreporting could lead to higher payments or even exclusion, reducing the bidder’s utility to zero. Thus, truthful bidding \( Bid_{b_s} = v_{b_s} \) is optimal, as it aligns the payment with 
the actual utility increment, maximizing the bidder's utility.

\subsection*{Supplementary Notes on Experiments}

\begin{figure}[h]
    \centering
    \hspace{-2.3em}
    \subfloat[mnist]
    {
        \includegraphics[width=0.49\linewidth, height=0.38\linewidth]{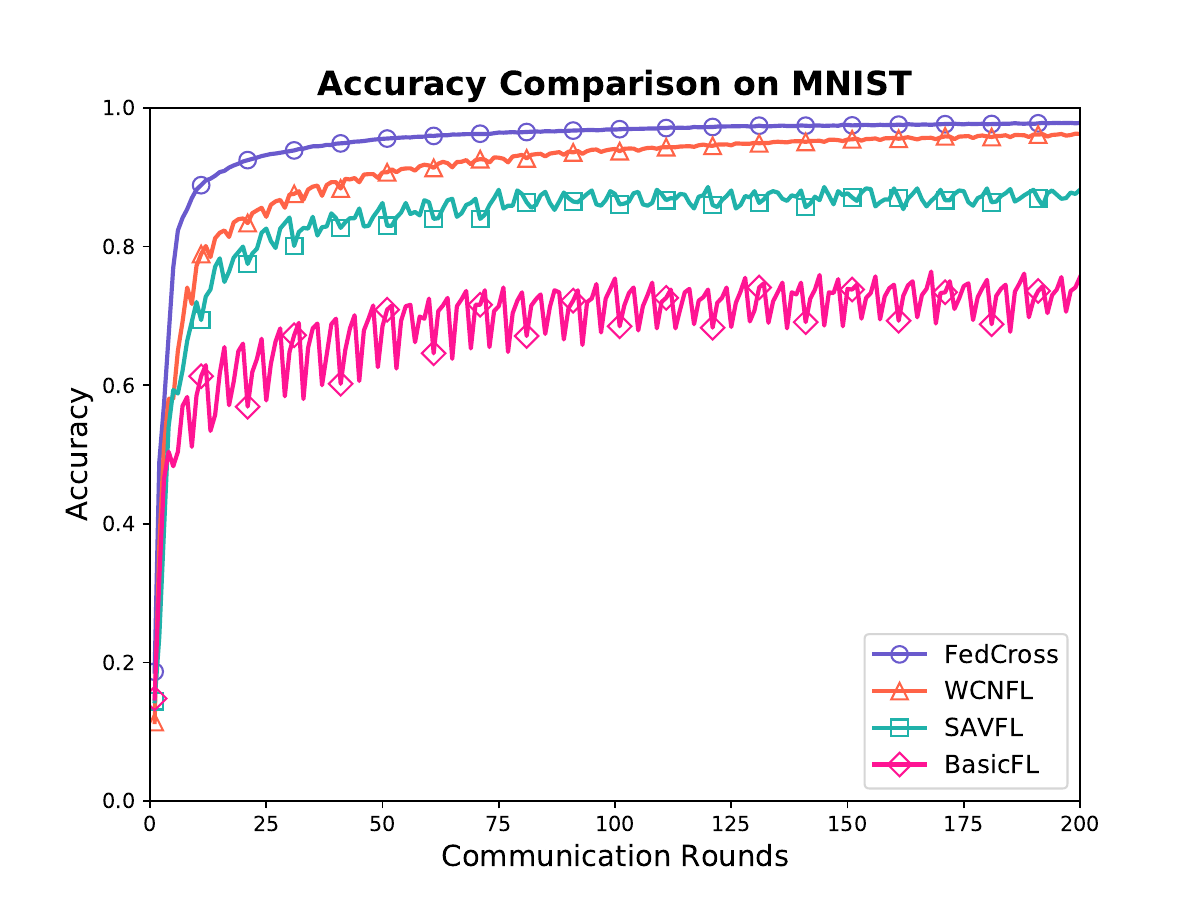}
        \label{fig:mnist}
    }
    \subfloat[cifar10]
    {
        \includegraphics[width=0.49\linewidth, height=0.38\linewidth]{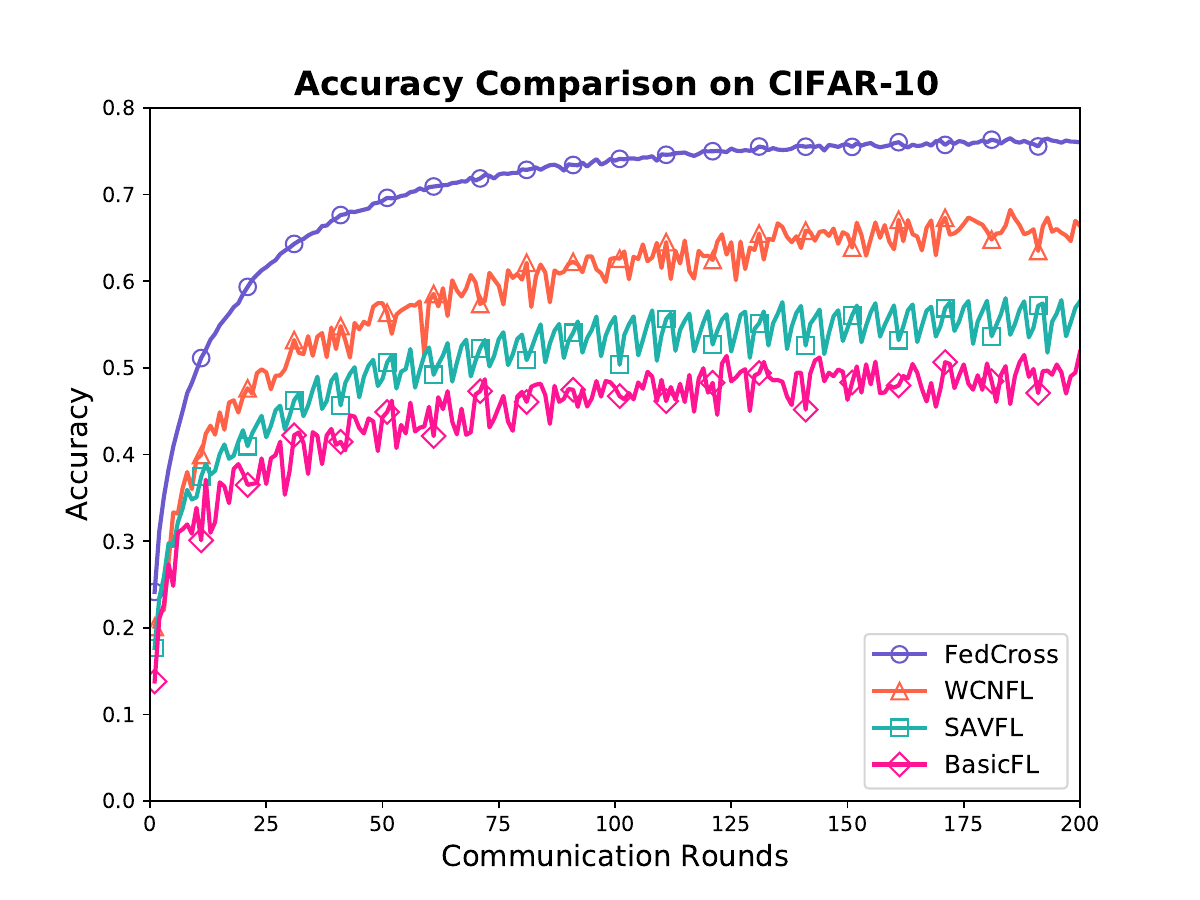}
        \label{fig:cifar10}
    }
    \caption{Accuracy of FedCross on Real Datasets}
    \label{fig:Fig_all_2}
\end{figure}

\textbf{Settings.} In this experiment, we used the PyTorch framework to implement FedCross and its baseline algorithm. All models were trained in the same hardware environment to ensure the fairness of the experimental results. We evaluate our method, FedCross, on real datasets, including MNIST \cite{Mnist} and CIFAR-10 \cite{Cifar10}, both of which were augmented with geospatial features \cite{Asynchronous-FL}. Specifically, we incorporated geographic information into the MNIST and CIFAR-10 datasets, allowing us to emulate a real-world scenario where geospatial data plays a role in federated learning tasks. MNIST consists of 70,000 images, and CIFAR-10 consists of 60,000 images with 10 classes. We used LeNet \cite{Mnist} for MNIST and a CNN for CIFAR-10 in the experiments.

\textbf{MNIST.} As a relatively simple dataset, MNIST allows most methods to achieve good results. FedCross, however, enables users to perform cross-region transfer tasks, ensuring the continuity of federated learning (FL) training. By continuously learning more effective strategies, it forms a more stable cluster, which leads to higher accuracy. WCNFL follows closely behind, benefiting from the incentive mechanism of reverse auctions, which motivates users. For SAVFL, the simulated annealing algorithm optimizes user migration target selection but does not account for the frequent mobility of users, which may result in suboptimal task allocation. Finally, the poor performance of BasicFL can be attributed to its simple framework, which struggles to adapt to the complexities of dynamic mobile networks, resulting in a less accurate model.

\textbf{CIFAR-10.} This dataset is more complex and requires more sophisticated models and longer training cycles to achieve better results. As a result, BasicFL's performance shows a more noticeable disadvantage compared to the MNIST dataset. WCNFL and SAVFL face similar challenges; while they improve task allocation efficiency to some extent, they still fail to fully address issues such as frequent interruptions and resource allocation in task migration. In contrast, FedCross's advantages in cross-region task migration and user strategy adjustment are more evident, allowing it to better handle complex datasets like CIFAR-10.

\end{document}